\documentclass[letterpaper, 10pt, conference]{ieeeconf}      %

\IEEEoverridecommandlockouts                              %

\usepackage{xspace}
\usepackage{colortbl} %
\usepackage{graphicx}
\usepackage{booktabs}
\usepackage{multirow} 
\usepackage[dvipsnames]{xcolor}
\usepackage{caption}
\usepackage{tabularx, microtype}
\usepackage{amsmath}
\usepackage{pifont}
\usepackage{amssymb} %
\usepackage{tablefootnote}
\usepackage{float}
\usepackage{makecell}
\usepackage[normalem]{ulem}
\def\eg{\textit{e.g.,}\xspace} 
\def\ie{\textit{i.e.,}\xspace} 
\def\etal{\textit{et al.}\xspace}

\def\task{Detection and Localization of Instruction Errors\xspace}
\newcolumntype{s}{>{\columncolor[gray]{.85}[.5\tabcolsep]}c}

\newcommand{\red}{BrickRed}
\newcommand{\green}{PineGreen}

\newcommand{\oursMethod}{IEDL\xspace}
\newcommand{\oursMethodName}{Instruction Error Detector \& Localizer\xspace}

\newcommand{\CLIPMethod}{CLIP Alignment\xspace}
\newcommand{\commonsense}{$^{\star}$\xspace}
\newcommand{\atd}{\texttt{ATD}\xspace}
\newcommand{\auc}{\texttt{AUC}\xspace}

\newcommand{\trajectoryset}{\Gamma}
\newcommand{\instructionTokens}{\Upsilon}
\newcommand{\datasetName}{R2RIE-CE\xspace}

\DeclareUnicodeCharacter{21BB}{$\circlearrowleft$}
\makeatletter
\let\NAT@parse\undefined
\makeatother

\usepackage{tikz}

\newcommand\copyrighttext{%
  \footnotesize \textcopyright 2024 IEEE. Personal use of this material is permitted. Permission from IEEE must be obtained for all other uses, including reprinting/republishing this material for advertising or promotional purposes, collecting new collected works for resale or redistribution to servers or lists, or reuse of any copyrighted component of this work in other works.}

\newcommand\copyrightnotice{%
\begin{tikzpicture}[remember picture,overlay]
\node[anchor=south,yshift=10pt] at (current page.south) {\fbox{\parbox{\dimexpr0.80\textwidth-\fboxsep-\fboxrule\relax}{\copyrighttext}}};
\end{tikzpicture}%
}

\title{\LARGE \bf Mind the Error! Detection and Localization of \\Instruction Errors in Vision-and-Language Navigation}

\author{Francesco Taioli$^{1,4}$, Stefano Rosa$^{2}$, Alberto Castellini$^{1}$, Lorenzo Natale$^{2}$,\\ Alessio Del Bue$^{2}$, Alessandro Farinelli$^{1}$, Marco Cristani$^{1}$, Yiming Wang$^{2,3}$
\\{\tt\small \url{https://intelligolabs.github.io/R2RIE-CE/}}
\thanks{$^{1}$ University of Verona, Verona, Italy.}
\thanks{$^{2}$ Istituto Italiano di Tecnologia (IIT), Genova, Italy.}
\thanks{$^{3}$ Fondazione Bruno Kessler, Trento, Italy. }
\thanks{$^{4}$ Polytechnic of Turin, Turin, Italy.\newline \phantom{--------}{\tt\small francesco.taioli@polito.it}}
\thanks{We acknowledge the CINECA award under the ISCRA initiative, for the availability of high-performance computing resources and support. This work was partially sponsored by the PNRR project FAIR - Future AI Research (PE00000013), under the NRRP MUR program funded by the NextGenerationEU. This project received funding from the European Union's Horizon Research and Innovation Programme G.A. n. 101070227. This study was also carried out within the PNRR research activities of the consortium iNEST (Interconnected North-Est Innovation Ecosystem) funded by the European Union Next-GenerationEU (Piano Nazionale di Ripresa e Resilienza (PNRR) – Missione 4 Componente 2, Investimento 1.5 – D.D. 1058  23/06/2022, ECS\_00000043)}%
}

\usepackage{hyperref}
\hypersetup{
    colorlinks=true,
    pdftitle={Mind the Error! Detection and Localization of Instruction Errors in Vision-and-Language Navigation},
    }

\begin{document}

\bstctlcite{IEEEexample:BSTcontrol}

\maketitle
\thispagestyle{empty}
\pagestyle{empty}
\copyrightnotice

\begin{abstract}
Vision-and-Language Navigation in Continuous Environments (VLN-CE) is one of the most intuitive yet challenging embodied AI tasks. 
Agents are tasked to navigate towards a target goal by executing a set of low-level actions, following a series of natural language instructions. All VLN-CE methods in the literature assume that language instructions are exact.
However, in practice, instructions given by humans can contain errors when describing a spatial environment due to inaccurate memory or confusion. Current VLN-CE benchmarks do not address this scenario, making the state-of-the-art methods in VLN-CE fragile in the presence of erroneous instructions from human users.
For the first time, we propose a novel benchmark dataset that introduces various types of instruction errors considering potential human causes. 
This benchmark provides valuable insight into the robustness of VLN systems in continuous environments. 
We observe a noticeable performance drop (up to $-25\%$) in Success Rate when evaluating the state-of-the-art VLN-CE methods on our benchmark.
Moreover, we formally define the task of Instruction Error Detection and Localization, and establish an evaluation protocol on top of our benchmark dataset. We also propose an effective method, based on a cross-modal transformer architecture, that achieves the best performance in error detection and localization, compared to baselines. 
Surprisingly, our proposed method has revealed errors in the validation set of the two commonly used datasets for VLN-CE, \ie R2R-CE and RxR-CE, demonstrating the utility of our technique in other tasks.

\end{abstract}

\section{Introduction}
\label{sec:introduction}

Interacting with agents through natural language is 
a long-term goal of embodied AI as it is potentially the most intuitive mode for human-robot communication.
The emerging research on Vision-and-Language Navigation~(VLN)~\cite{Anderson_2018_vln,gu2022vision} is along this path, aiming to develop embodied agents that, following a given instruction in the format of natural language, can reach a target destination in a 3D environment, \eg \textit{``Exit the bedroom and turn left. Walk straight past the grey couch and stop near the rug."}
VLN is a challenging task. First, it requires a \emph{strong} vision-language alignment, \ie the alignment between the visual information (RGB-D inputs) and the natural language instruction. Second, VLN agents need to reason about which part of the instruction has been executed, and which parts need to be carried out.
Lastly, agents must be able to generalize from seen to unseen environments, as well as from simulation to real-world~\cite{habitat_19_iccv}.

\begin{figure}[!t]
    \centering
    \includegraphics[width=1\linewidth ,trim=3 3 3 3,clip]{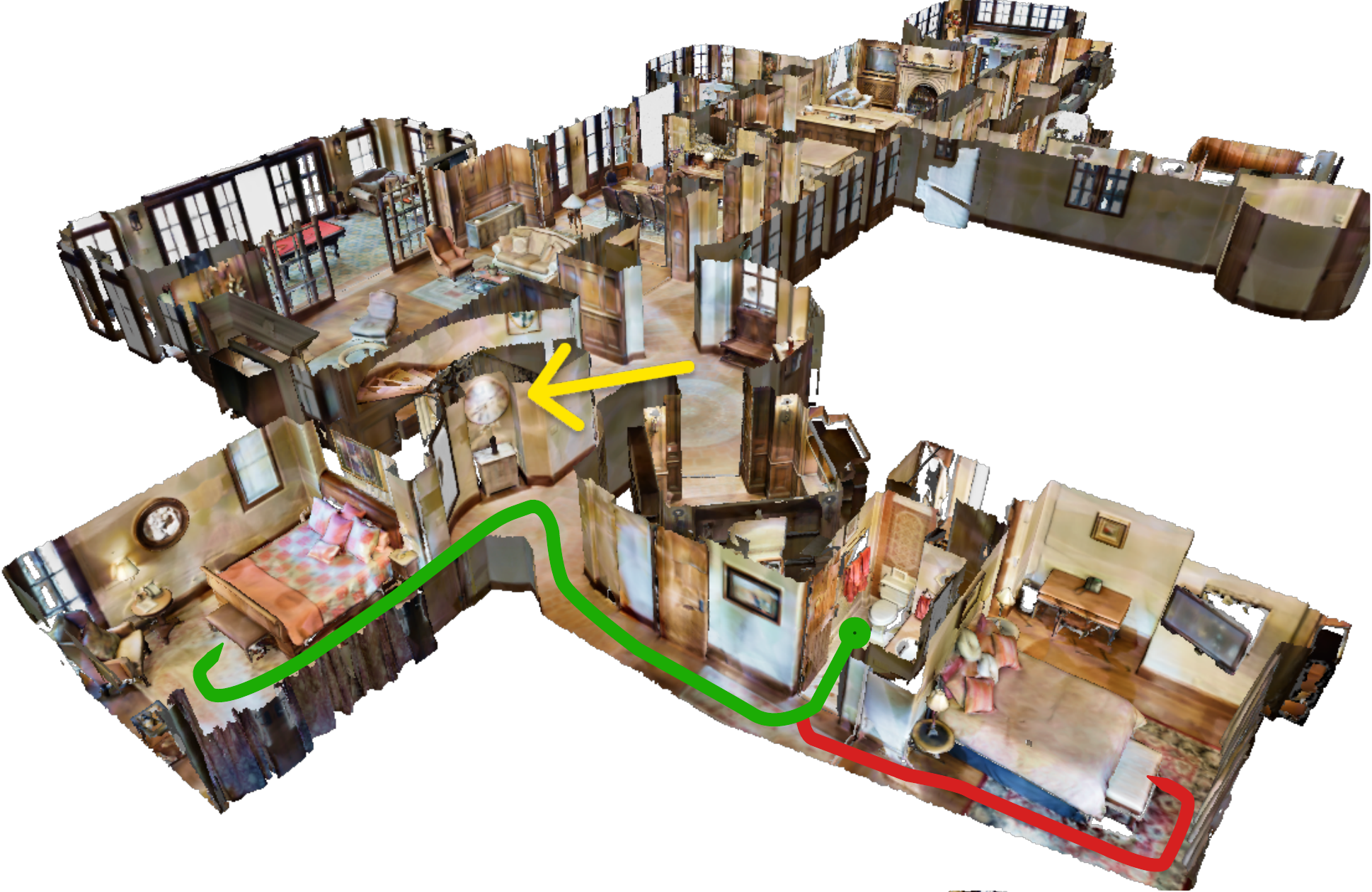}
    \caption{An agent navigates in a scene, following instructions expressed in natural language, for example \textit{``Exit the bathroom and go {\color{\red}left} ({\color{\green}\checkmark right}), then turn left at the \underline{big clock} and go into the bedroom and wait next to the bed."} By just changing ``right" to ``left" in the instruction, the agent terminates the exploration in the wrong location, ignoring the fact that along the path it did not see the ``big clock" (yellow arrow).}
    \label{fig:teaser}
    \vspace{-6mm}
\end{figure}

\begin{figure}[t]
    \centering
    
    \includegraphics[width=1\linewidth]{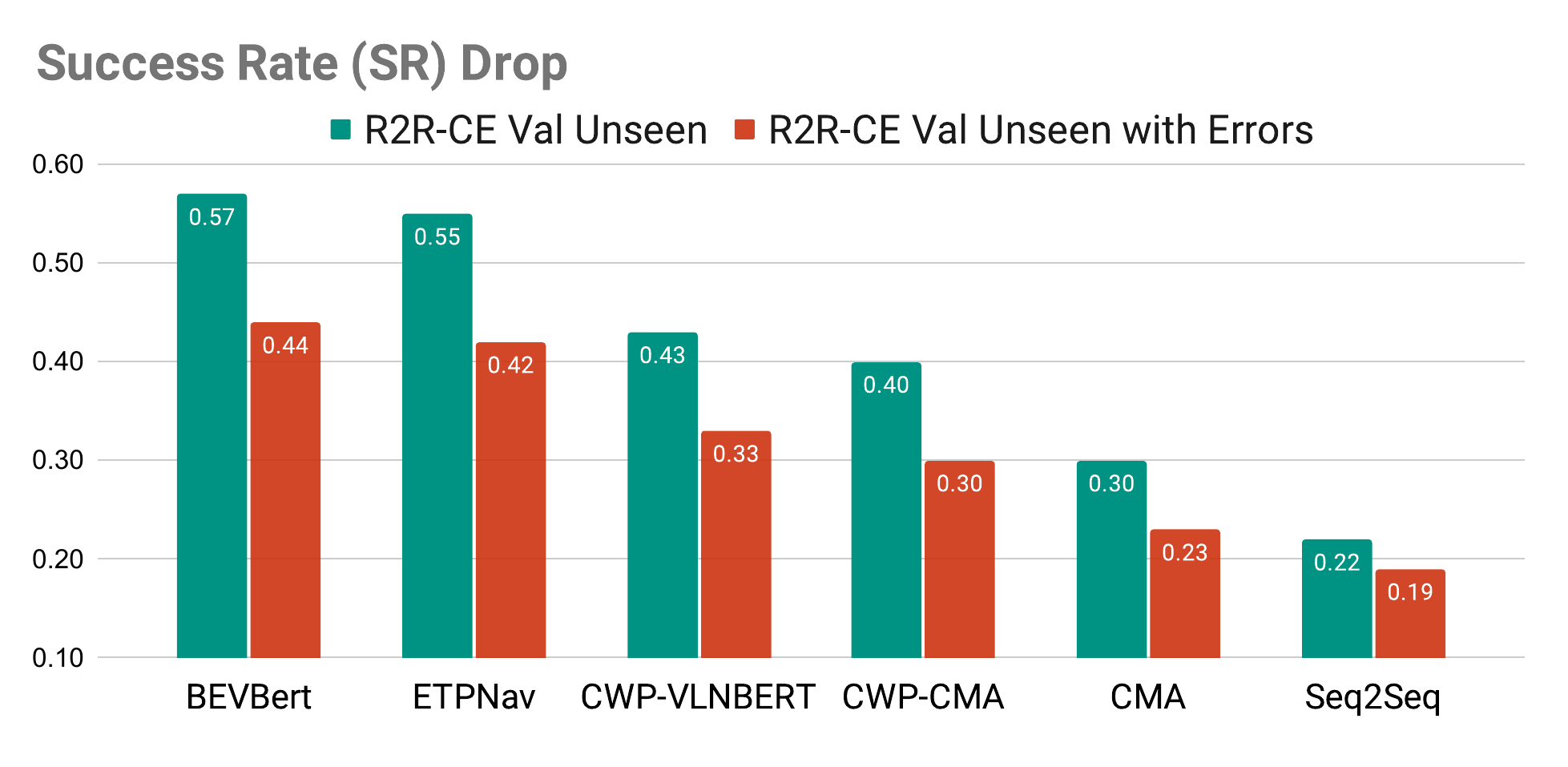}
    \caption{Comparison of the Success Rate (\texttt{SR}) of different methods (in order, \cite{an2023bevbert,an2023etpnav,discrete_to_cont,discrete_to_cont,krantz_vlnce_2020,krantz_vlnce_2020}) working on continuous environments\footref{fn:sota_tab1}. We show the \texttt{SR} on the standard R2R-CE dataset split \texttt{Val Unseen} (green) and the drop in \texttt{SR} performance when errors are present (red). Interestingly, we see up to $-25\%$ drop in \texttt{SR} when up to three errors among\textit{ \{Direction, Room, Object\}} per episode are present.}
    \label{fig:sr_drop}
    \vspace{-10pt}
\end{figure}

To facilitate the study of VLN, many benchmark datasets have been proposed. For instance, Room-to-Room (R2R)~\cite{Anderson_2018_vln} is the first benchmark dataset for VLN, which is built on top of Matterport3D~\cite{matterport_dataset} scenes and it is equipped with discrete navigation graphs (\ie discrete environments)~\cite{Anderson_2018_vln}. Follow-up works have further extended the R2R dataset from different aspects, \eg enabling VLN in the continuous environments, a more practical scenario, where agents can navigate to any unobstructed point via a set of low-level actions (R2R-CE)~\cite{krantz_vlnce_2020}, levelling up the scale of the dataset
~\cite{wang2023scalevln}, or providing multilingual fine-grained visual groundings that relate each word to pixels/surfaces in the environment~\cite{rxr}.
All the previous benchmarks, however, only consider \emph{correct} language instructions. This consideration can be brittle in reality as human often gives instructions that are approximate or ambiguous, or even prone to error as based on their memory. %
Moreover, in some cases, subjects with cognitive or perceptual impairments may provide inaccurate descriptions when naming a task~\cite{spatial_nav_def}.
As shown in Fig.~\ref{fig:teaser}, simply changing the word \textit{``right"} to \textit{``left"} can cause the agent to terminate the navigation in a wrong location, even if the central part of the instruction still contains informative details to discern that the sentence has some mistakes.
Our preliminary studies, summarized in Fig.~\ref{fig:sr_drop}, show that state-of-the-art VLN methods\footnote{Leaderboard at \href{https://eval.ai/web/challenges/challenge-page/719/leaderboard/1966/success}{https://eval.ai/web/challenges/challenge-page/719/leaderboard/1966/success}. Only models with public weights are evaluated, on the latest version of the R2R-CE dataset (version 1.3)\label{fn:sota_tab1}}
fed with \textit{(i)} correct instructions (in green) and \textit{(ii)} instructions with up to $3$ errors (in red) have a noticeable gap (up to -$25\%$) in terms of Success Rate.
This analysis shows that it is critical to understand the presence of errors in the instruction and to locate them. 

In this work, we first formally define the types of errors that may occur in language instructions for the VLN task in indoor environments, including \textit{Direction, Room, Object, Room\&Object} and a combination of \textit{All} types of errors. Based on these definitions, we propose a novel benchmark in continuous environments built on top of the R2R-CE dataset~\cite{krantz_vlnce_2020}, in which we artificially inject errors of the different types to thoroughly evaluate  
 the capability of VLN methods to deal with these mistakes. %
 Moreover, we propose a novel task, \ie \emph{\task}, which aims to detect and localize errors within a given instruction, an important intermediate task for further addressing instruction errors in VLN policy learning.
 Then, we propose a method based on a cross-modal transformer, fusing together the language features of the instruction with the observations of the agent, achieving competitive performance in solving the \task, compared to a \CLIPMethod baseline. 
Notably, our approach could be useful in spotting annotation errors in existing benchmarks: we discovered 8 episodes in the R2R-CE and 10 episodes in the RxR-CE dataset that contain either incorrect or ambiguous instructions, which should not be considered during evaluation.
Our contributions are summarized below:
\begin{itemize}
    \item We categorize the errors that exist in language instructions in the VLN-CE task, and establish the first benchmark (\datasetName) for VLN in continuous environments with instruction errors.

    \item We experimentally show that state-of-the-art VLN methods are not robust to instruction errors using our proposed benchmark, necessitating the study of instruction errors in VLN.
    
    \item We formalize the novel task of \textit{\task} for VLN agents, and propose an effective method, \oursMethodName (\textit{\oursMethod}), based on a novel Instruction-Trajectory compatibility model. Our method effectively connects the semantic meaning of the language instruction with the sequence of visual observations of the agent, establishing a competitive baseline.

    \item We discover errors in the ground-truth annotations of the R2R-CE dataset, and episodes that should be removed from the validation split of RxR-CE dataset, demonstrating an additional use of our technique.
\end{itemize}

\section{Related Works}

\label{sec:related}
We discuss recent methods on VLN and failure analysis on VLN methods. We also cover the relevant study on instruction and trajectory alignment.

\textbf{Vision-and-Language Navigation.}
The VLN task~\cite{Anderson_2018_vln} has drawn increasing attention since its introduction. The seminar work~\cite{Anderson_2018_vln} presents the Room-to-Room (R2R) navigation task and the Matterport3D Simulator
based on the Matterport3D indoor environment~\cite{matterport_dataset} with real 360-degree RGB-D scans. Importantly, methods based on the R2R navigation task define their action space based on an undirected graph that is constructed from panoramic viewpoints (\ie nav-graph), forming a discrete environment. Early VLN agents operating under this paradigm capture historical observation by means of recurrent unit~\cite{Anderson_2018_vln,Wang_2019_CVPR,speaker_follower} or by using attention mechanism~\cite{hamt,recurrent_vln_bert,duet_cvpr}.
Instead, in~\cite{krantz_vlnce_2020}, VLN in Continuous Environment (VLN-CE) is introduced, in which agents are allowed to move freely, thus removing the assumption of known environment topologies, short-range oracle navigation and perfect agent localization. In doing that, they translated the nav-graph R2R trajectories to the continuous environments in the Habitat simulator~\cite{habitat_19_iccv}. The authors find significantly lower absolute performance in the continuous settings.
To bridge the discrete-continuous gap,~\cite{discrete_to_cont} propose a predictor to generate a set of candidate waypoints during navigation. Given the long time horizon of an episode under this setting, recent methods proposed metric maps~\cite{grid_mm} or topology memory~\cite{an2023etpnav,an2023bevbert} to represent the history of observations. 
The current state-of-the-art method, BEVBert~\cite{an2023bevbert}, proposed a hybrid map to balance the demand for both short-term reasoning and long-term planning while introducing a pre-training map paradigm. 
Differently from the above-mentioned VLN works, in which the focus is to reach the target goal in the most efficient way, we aim at introducing and addressing a novel task, namely the \emph{\task} within the VLN context.

\textbf{Failure Analysis on VLN.}
Several works have investigated the behaviour and failure of VLN agents. The study in~\cite{zhu2022diagnosing} shows that the agents refer to both object and direction tokens when making their navigation decisions.
A similar work~\cite{limitation_vln_agent_aamas} further evaluates to which degree the spatial and directional language cues inform the navigation decisions. They focus on the path-ranking models and replace directions/nouns/objects and numbers tokens with the \texttt{[MASK]} token.
The study in~\cite{diagnosin_env_bias} investigates the significant drop in performance when tested on unseen environments, locating the environment bias to be in the low-level visual appearance. Moreover, Yang \etal~\cite{Behavioral_Analysis_Yang_2023_CVPR} developed a methodology to study agent behaviour on a skill-specific basis. Their method is based on generating skill-specific interventions and measuring changes in action predictions, considering the stop behaviour, unconditional (directional) and conditional (object- and room-seeking) skills.

Our work differs from the previous studies, as we investigate the instruction errors and their impact for navigation policies in the VLN task. We manipulate among a large set of tokens to inject instruction errors, instead of just masking certain tokens out as in~\cite{limitation_vln_agent_aamas}. We do not focus on path-ranking models~\cite{diagnosin_env_bias}, the domain gap towards unseen environments~\cite{diagnosin_env_bias}, and skill-specific interventions~\cite{Behavioral_Analysis_Yang_2023_CVPR}.

\textbf{Instruction and Trajectory Alignment.}
Instruction and Trajectory Alignment is the ability to align semantically the concepts represented by these two modalities, an important aspect for obtaining a VLN policy. Huang \etal~\cite{Huang_2019_ICCV} introduces the Cross-Modal Alignment (CMA) task, \ie discriminating instruction-trajectory path pairs from negative pairs, by maintaining the original instruction and altering the path sequence.
Zhao \etal~\cite{on_the_evaluation_of} perform a study on the quality of the instruction generated by VLN instruction generation models~\cite{speaker_follower,env_drop} using human wayfinders. Moreover, they propose an Instruction-Trajectory compatibility model to classify high and low-quality instructions for trajectories from the R2R \texttt{Val Unseen} and \texttt{Val Seen} sets with the discrete VLN setting. These instructions are considered to be high quality if 2 out of 3 human wayfinders reached the goal.
Differently from~\cite{Huang_2019_ICCV}, they also consider instruction perturbations.
In~\cite{Liang_aaai_contrastive} the authors proposed a Contrastive Instruction-Trajectory Learning framework to boost the generalizability of the navigation policy. Their method selects sub-optimal trajectories by the hop distance from an anchor trajectory. Positive instructions are generated by substituting words with their synonym, inserting or substituting them with context or by back-translation; instead, negative ones are generated by splitting the original instruction into sub-instructions and shuffling or repeating randomly to become an intra-negative instruction. 
The authors assume that the augmented instructions should preserve the semantic information of the original ones.

While the above-discussed works are relevant, their objective is to obtain a better Instruction and Trajectory Alignment for the VLN policy learning. Instead, our work investigates how instruction errors impact the VLN policy, and addresses the novel task of detection and localization of instruction errors for a given policy. Moreover, our study is based on VLN-CE, while the Instruction and Trajectory Alignment is mostly based on the discrete VLN operating on top of the nav-graph.

\section{Task Definition}
\label{sec:task_definition}
In this section, we first introduce VLN in Continuous Environments (VLN-CE); then, we define the types of instruction errors and finally formalize our novel task of \textit{\task}.

\textbf{VLN-CE.}
For each episode, the agent is given an instruction $\mathcal{I}$ composed of $F$ words, \ie $\mathcal{I} = \{w_1, ..., w_F\}$.
At each time step $t$, the agent receives a visual observation $O_t$, \ie an RGB-Depth image, collected during the navigation.
We define $\mathcal{O}=\{O_1,..., O_T\}$ as the set of visual observations over a navigation episode, where $T$ is the total number of steps executed into the environment by a policy $\pi$. 
The goal of VLN-CE is to learn a policy $\pi$ for reaching a target goal, which maps the instruction $\mathcal{I}$ and observation $O_t$ to a set of low-level actions $a_t$ $\in \{$\texttt{Forward 0.25m, Turn Left 15°, Turn Right 15°, Stop} $\}$.  Note that in this work we assume the policy $\pi$ to be given, \eg the state-of-the-art policy~\cite{an2023bevbert}, since we are interested in solving the task of \task, not the VLN-CE task.

\textbf{Types of Instruction Error.}
Humans are prone to errors when providing directions to a destination. For instance, one might say,  \textit{``Go down the hallway and turn {\color{\red}right} ({\color{\green}\checkmark left}) when you see the plant near the {\color{\red}bathroom} ({\color{\green}\checkmark bedroom})".}
We categorize the common instruction errors based on their semantics, to facilitate a systematic assessment of their impact on VLN-CE. Furthermore, for a better practical significance, we focus on instruction errors that are likely to occur in the real world, considering common human mistakes due to either confusion or inaccurate memory of a scene structure.
Specifically, we identify three types of individual errors, detailed below:

\textit{(i)~Direction Error}: words indicating directions, such as \textit{left} or \textit{backward}, are important ingredients in human instructions for navigation tasks. 
Each direction has its natural antonym counterpart, such as \textit{left/right}, or \textit{forward/backward}, which can contribute to common instruction errors due to human confusion. If an agent is wrongly given an instruction with the opposite direction, \eg~{\color{\red} left} ({\color{\green}\checkmark right}), its path can deviate heavily from the correct one, as illustrated in Fig.~\ref{fig:teaser}. The occurrence of a \emph{direction error} is defined if at least one ground-truth direction is replaced by a different direction word in an instruction. We mainly focus on the more frequent \emph{direction error}, in which the antonym direction replaces the ground-truth direction.

\textit{(ii)~Object Error}: words indicating objects that are observed along the navigation also play an important role in guiding agents' motion. However, humans may not remember the exact scene including the object's disposition. Humans could likely confuse objects in the instruction, especially among those that are often co-located in a common space. For example, an error such as {\color{\red} sofa} ({\color{\green}\checkmark chair}) is more likely to happen than {\color{\red} toilet} ({\color{\green}\checkmark chair}). We define an \emph{object error} as an error that occurs if at least one object class in the instruction is wrongly indicated to a different type. In particular, we focus on \emph{object errors} that account for such \emph{common sense} priors, \ie object co-occurrence, in a common space.

\textit{(iii)~Room Error}: words indicating rooms are fundamental to guide the agent towards its destination since they provide context along the navigation. As with \emph{object errors}, humans may confuse one room with another when providing instructions, or they may not recall the exact layout of the scene. It is more likely humans confuse rooms among those that are more geometrically adjacent to each other. For example, as the bedroom is often next to the bathroom, the instruction likely contains the error as: \textit{``Go into the door. Once inside, go into the {\color{\red}bathroom} ({\color{\green}\checkmark bedroom}). Stop in front of the closet, near the bed."}. 
We can define the \emph{room error} as the occurrence of at least one room in the instruction is wrongly indicated to another room type. In particular, we focus on \emph{room error} that accounts for the room adjacency priors. 

\textit{(iv)~Room\&Object Error}: as both \emph{object errors} and \emph{room errors} are attributed to the inaccurate human memory regarding a scene, we also consider the \emph{Room\&Object errors} as cases in which both types of errors co-occur in a given instruction.
For example,~\textit{``Exit the room and turn right. Continue forward until you arrive at the {\color{\red}living room} ({\color{\green}\checkmark kitchen}). Go around the table and stop in front of the {\color{\red}lamp} ({\color{\green}\checkmark sink}),  near the stove."}

\textit{(v)~All Error}: finally, all individual types of errors can occur in the same instruction due to inaccurate scene-memory and direction confusion. An \emph{all error} exists if there is at least a \emph{direction error} and a \emph{room\&object error}. For example,~\textit{``Exit the bathroom and go {\color{\red}left} ({\color{\green}\checkmark right}), then turn left at the big clock and go into the {\color{\red}bathroom} ({\color{\green}\checkmark bedroom}) and wait next to the {\color{\red}closet} ({\color{\green}\checkmark bed}) in front of the armchair."}.

\textbf{Instruction Error Detection and Localization.}
Given an instruction $\mathcal{I}$, which describes in natural language how to reach a target position, and a set of observation $\mathcal{O}$ collected by a VLN agent controlled by a given policy $\pi$, the task of \textit{Instruction Error Detection} aims to learn a function
\mbox{$d_{\pi}: \mathcal{I} \times \mathcal{O} \rightarrow \{ \text{True}, \text{False} \}$}, that returns True if the instruction contains errors, False otherwise. 
If the detection function $d_{\pi}$ returns True, the task of \textit{Instruction Error Localization} can also be performed. It aims to identify the positions of the error occurrences within the instruction. Namely, positions of wrong words in the instruction are computed. Formally, the localization function can be defined as~\mbox{$l_{\pi}:\mathcal{I} \times \mathcal{O} \rightarrow \mathcal{P}( \{0, 1, \ldots, len(\text{$\mathcal{I}$}) - 1\})$}, where $len(\mathcal{I})$ is the total number of words in the instruction~$\mathcal{I}$, and $\mathcal{P}(\cdot)$ is the power set operator.

\section{Benchmark}
\label{sec:benchmark}
We establish the first benchmark dataset \textbf{\datasetName}, short for \textit{R2R with Instruction Errors in Continuous Environments}, for assessing VLN-CE methods with erroneous instructions and for the novel task of \task. 
Our benchmark dataset is built on top of the R2R-CE~\cite{krantz_vlnce_2020} dataset, which serves for evaluating VLN-CE methods. 
The R2R-CE dataset is split into \texttt{Train}, \texttt{Val Seen} and \texttt{Val Unseen}. We will use \texttt{Train} set for training data and focus on the most challenging \texttt{Val Unseen} for validation, which has $1839$ episodes with new paths, instructions, and scenes that are not observed during training.

In order to create erroneous instructions, we artificially inject various types of Instruction Errors (as defined in Sec.~\ref{sec:task_definition}) into the given natural language instructions in R2R-CE. Our error composition carefully considers error priors induced by human causes, including inaccurate scene memory and direction confusion. For each type of error, \ie \textit{Direction}, \textit{Object}, \textit{Room}, \textit{Room\&Object} and \textit{All}, we create a corresponding validation set based on the \texttt{Val Unseen} validation split of R2R-CE. 
The original validation split can contain episodes that may not be appropriate for our evaluation. The natural language instruction should not be too short and the episodes should contain words relevant to the specific error setup. For example, episodes must have Direction words for the validation set with \textit{Direction Error}, or episodes must contain Direction, Object, and Room words for the validation set with \textit{All Errors}.
Specifically, prior to constructing the validation set with each type of error, we first exclude episodes with a minimum length of less than $\tau$ words, and episodes lacking words relevant to the specific experiments, such as Direction error words for the \textit{Direction Error} dataset.
Following this filtering stage, we obtain a set of correct episodes $\mathcal{E_{C}}$ for each type of Instruction Error. 

Then, for each episode $e_i \in \mathcal{E_{C}}$, we create a corresponding erroneous episode in which we perturbed the instruction with the respective error. 
Specifically, 
\textit{(i)} for the \textit{Direction Error}, we consider the following set of directional words that occur frequently in the instructions: \textit{left/right}, \textit{go down/go up}, \textit{into/out of}, \textit{forward/backward}, \textit{inside/outside}, \textit{go around/go back}, \textit{leftmost/rightmost}. We introduce the \textit{Direction Error} by swapping the ground-truth direction with its antonym;
\textit{(ii)} For the \textit{Object Error}, we first identify a set of common object categories $\mathcal{C}$ that frequently appear in a set of language instructions, excluding synonyms. Each object class $c_i \in \mathcal{C}$ is associated with a set of object classes $\mathcal{C}_i$ comprised of the classes that often co-locate in the same room. We introduce an \textit{Object Error} by swapping the ground-truth object class $c_i$ to a random class $c_j \in \mathcal{C}_i$;
\textit{(iii)} For the \textit{Room Error}, we consider the following set of rooms $\mathcal{R}$ that are common in indoor environments: \textit{kitchen, archway, bathroom, bedroom, gym, lounge, hallway, living room, office, dining room, laundry, restroom}. Each room $r_i \in \mathcal{R}$ is associated with a set of rooms $\mathcal{R}_i$ that are often adjacent to $r_i$\footnote{The list is obtained by leveraging ConceptNet relations and manual curation}. We introduce a \textit{Room Error} by swapping the ground-truth room $r_i$ to a random room $r_j \in \mathcal{R}_i$;
\textit{(iv)} For the \textit{Room\&Object Error}, we introduce in an instruction both \textit{Room} and \textit{Object Error}; 
\textit{(v)} For the \textit{All Error}, we introduce in an instruction both \textit{Direction} and \textit{Room\&Object Error}. 
The perturbed episodes will then be stored in the corresponding set of perturbed episodes, $\mathcal{E_{P}}$. 
For each perturbed episode $e_i \in \mathcal{E_{P}}$, both the error type and the position of the perturbed word are stored as metadata for the assessment of \task. 
Eventually, for each type of instruction error, we obtain sets $\mathcal{E_{C}}$ and $\mathcal{E_{P}}$. Tab.\ref{tab:statistic} presents the statistics of the created validation set with each type of error.

\vspace{-0.3cm}
\begin{table}[tbh]
\centering
\caption{Statistics of the dataset used in this benchmark.}
\label{tab:wrong_ep_removal}
\resizebox{0.8\columnwidth}{!}{
    \begin{tabular}{cccc}
    \hline
    Error type & \# Episodes & \makecell{Errors per\\episode} & \makecell{Mean instr.\\ length (tokens)} \\
    \hline
    Direction       & 3218 & 1 &  31.59 \\
    Room            & 2064 & 1 & 32.05 \\
    Object          & 3162 & 1 & 30.94 \\
    
    Room \& Object  & 1734 & 2 & 33.35 \\
    
    All             & 1586 & 3 & 34.58 \\
    \hline
    \label{tab:statistic}
    \end{tabular}}
    \end{table}

\vspace{-0.5cm}
\section{Method}
\begin{figure*}[t]
    \centering
    \includegraphics[width=0.9\linewidth]{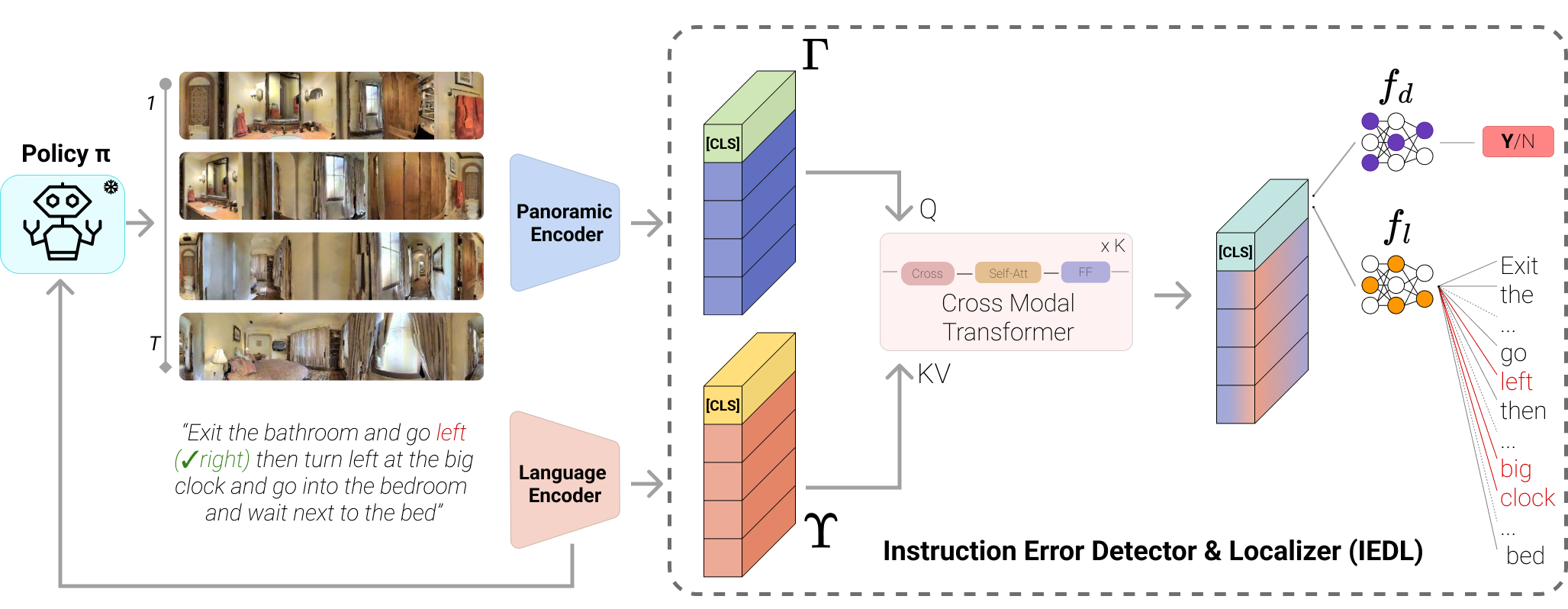}
    \caption{Architecture of our proposed \textit{\oursMethod} model, representing the scenario depicted in Fig~\ref{fig:teaser}. The frozen policy $\pi$ follows Instruction $\instructionTokens$, producing a sequence of observation $\mathcal{O}$. Then, a panoramic encoder and a language encoder produce, respectively, the trajectory visual features $\trajectoryset$ and instruction features $\instructionTokens$. We then feed the trajectory set $\trajectoryset$ and $\instructionTokens$ to a cross-modal multi-layer transformer to produce visual-language aligned features. Finally, two specialized heads perform \textit{Instruction Error Detection} and \textit{Instruction Error Localization}, respectively.}
    \label{fig:architecture}
    \vspace{-4mm}
\end{figure*}

We instantiate our \oursMethodName  (\emph{\oursMethod}) on BEVBert~\cite{an2023bevbert}, currently the state-of-the-art method for VLN-CE\footref{fn:sota_tab1}, but our method is \textit{model-agnositic}. BEVBert is composed of a hybrid map to balance the demand of VLN for both short-term reasoning and long-term planning:~\textit{(i)} a local metric map is used to explicitly aggregate incomplete observations and remove duplicates; \textit{(ii)} a \textit{graph-based} topological map equipped with a global action space is used for long-term planning, and \textit{(iii)} a pre-training framework, based on the hybrid map, is used to learn multimodal map representation which enhances spatial-aware cross-modal reasoning.
In the following, we will refer to policy $\pi$ as an optimal policy trained using the BEVBert model. Note that during the training process of our method, policy $\pi$ is frozen, thus it will not receive any parameter updates. %

\textbf{Model structure.} The full pipeline of \emph{\oursMethod} is shown in Fig.~\ref{fig:architecture}.
We define the instruction embeddings  as~$\instructionTokens\in \mathbb{R}^{W \times D}$, namely the textual instruction $\mathcal{I}$ undergoes a tokenization and padding procedure, up to $W=80$ tokens~\cite{an2023bevbert,an2023etpnav}, with associated text BERT~\cite{bert} embeddings. 
Moreover, we refer to $D$ as the dimensionality of the latent space.
We consider the trajectory set of image embeddings $\trajectoryset=\{V_1,..., V_T\}$ (where $V_t$ is the corresponding image embedding of visual observation $O_t$ defined in Section \ref{sec:task_definition}) as the set containing the viewpoint panorama visual features $V_t \in \mathbb{R}^D$ (\ie embeddings) of the node chosen at each time step $t$ by policy $\pi$, following instruction embeddings $\instructionTokens$. 
 As in~\cite{an2023bevbert}, the embedding $V_t$ contains features extracted from a pre-trained Vision Transformer (ViT-B/16-CLIP~\cite{clip}), which are then fed into a panoramic encoder~\cite{hamt} to obtain contextual view embeddings. 
For each episode $i$, we first collect the set of observation embeddings (\ie trajectory) $\trajectoryset \in \mathbb{R}^{T \times D}$ from policy $\pi$. To retain positional information, we use sine and cosine positional encoding.
Similar to BERT~\cite{bert} \texttt{[CLS]} token, we prepend a learnable \texttt{[CLS]} $\in \mathbb{R}^D$ embedding to set $\trajectoryset$, which will be later used for our downstream tasks. 
Then, trajectory set $\trajectoryset$ and instruction embeddings $\instructionTokens$ are fused together by feeding them to a cross-modal multi-layer transformer with $k$ layers (see right part of  Fig.\ref{fig:architecture}). 
We adopted an architecture similar to~\cite{tan-bansal-2019-lxmert} but without the bi-directional cross-attention layer since our focus is not on having more than one cross-modal aligned output. 
For each layer of the transformer, we first perform cross-attention using trajectory set $\trajectoryset$ as a query ($Q$) and instruction embeddings $\instructionTokens$ as a context ($KV$). 
Then, we perform self-attention followed by a feed-forward layer. 
The \texttt{[CLS]} token (enriched with visual-language fused features after the cross-modal transformer) is then fed to two classification heads that produce the output of our \oursMethodName:~\textit{(i)} the trajectory-instruction matching head $f_{d}: \mathbb{R}^{D}\rightarrow  \mathbb{R}$ (which implements function $d_{\pi}$ presented in Sec.~\ref{sec:task_definition}) predicts the alignment score $\sigma(a) \in [0,1]$, where $\sigma(\cdot)$ is the sigmoid operator;
\textit{(ii)} the error localization head $f_{l}: \mathbb{R}^{D}\rightarrow  \mathbb{R}^{W}$ (which implements function $l_{\pi}$ presented in Sec.~\ref{sec:task_definition}) predicts the token errors' localization within the instruction. Both heads are implemented by a Multilayer Perceptron (MLP), a ReLU activation function, followed by a Layernorm layer and a final MLP.

\textbf{Training.} 
We train the model with two separate losses. Specifically, $\mathcal{L}_{d}$ for the trajectory-instruction matching $f_{d}$ head is trained using Binary Cross-Entropy loss.
Instead, $\mathcal{L}_{l}$ for the error localization $f_{l}$ head is trained using standard Cross-Entropy Loss. Since multiple errors can be present within an instruction, we sum the loss for each localization term. Thus, we minimize the following loss $\mathcal{L}$:

\[\mathcal{L} = \lambda_1 \mathcal{L}_{d} + \frac{\lambda_2}{E}\sum_{i=1}^{E} \mathcal{L}_{l}\]
where $E$ is the number of errors actually present in the instruction, $\lambda_1$ and $\lambda_2$ are the hyperparameters for the weights of the two losses, respectively.

\section{Experiments}
We first benchmark a state-of-the-art policy~\cite{an2023bevbert} on our dataset, to show how different types of error affect its performance. We then evaluate the proposed \textit{\oursMethod} model on the \textit{\task} task in comparison to two baseline methods. Finally, we show \textit{\oursMethod} can identify episodes with annotation errors
in existing dataset. In the following, we introduce the performance metrics and the compared baselines. 

\textbf{Performance Metrics.} 
We use standard metrics for evaluating VLN performance as in prior works~\cite{Anderson_2018_vln,anderson2018evaluation}. An episode in the VLN task is considered successful if the distance between the final position of the agent and the target location is less than 3 meters. The main VLN metrics are Success Rate (\texttt{SR}), and Success rate weighted by (normalized inverse) Path Length (\texttt{SPL}).
Moreover, to assess the performance of the task of \textit{\task}, we adopt the Area Under the ROC Curve (\texttt{AUC}) as in~\cite{on_the_evaluation_of}, \ie the area beneath the True Positive Rate plotted against the False Positive Rate. We consider an instruction to be positive if it contains at least one error.
In this work, we consider \auc as the main metric.
To measure the performance regarding the task of \textit{Instruction Error Localization}, we introduce a novel measure termed as \emph{Absolute Token Distance} \texttt{(ATD)}. 
The \texttt{ATD} metric is defined as the absolute difference between the predicted position of the perturbed token and the true position of the perturbed token. 
Formally, for an episode $i$, let $\ell_j^i$ be the $j^{th}$ ground-truth index of the token that has been perturbed to form an error, and let $\hat{\ell}_j^i$ be the predicted index of the token in the instruction containing the error,
we define \texttt{ATD} as:
\begin{equation}
    ATD = \frac{1}{N_{\mathcal{E}_P}}\sum_{i=1}^{N_{\mathcal{E}_P}} \frac{1}{J_i}\sum_{j=1}^{J_i} |\ell_j^i - \hat{\ell}_j^i|,
\end{equation}
where $N_{\mathcal{E}_P}$ is the number of perturbed episodes $\mathcal{E}_P$, and $J_i$ is the number of perturbed tokens in the episode $i$.

\textbf{Implementation details.}
For details on BEVBert, we refer the reader to~\cite{an2023bevbert}. We set $k=4$ as the number of transformer layers for our multi-layer cross-modal transformer and $D=768$ for all the feature dimensions.
In our experimental trials, we set both $\lambda_1$ and $\lambda_2$ to 1 empirically for our loss function $\mathcal{L}$.
We train \emph{\oursMethod} model for up to 9k iterations, using the AdamW optimizer. 
Specifically, for all experiments described below, we train our \textit{\oursMethod} method using the \textit{All Error} type benchmark \textit{without} common sense from the \textit{training set} of R2R-CE, thus we do not enforce the co-location of objects from object set $\mathcal{C}$ and adjacency for room set $\mathcal{R}$. This ensures that additional common sense bias is not injected during training.
We then select the model that achieves the best \texttt{AUC} score, using the \textit{All} error type benchmark \textit{with} common sense from the validation splits \texttt{Val Unseen}.
The selected model is used for evaluation across the remaining benchmarks. 
 
\textbf{Baselines.}
As no baseline for this task is available, we compare \textit{\oursMethod} method with a random baseline and a simple zero-shot alignment baseline:

\noindent\textit{(i) Random}. Each instruction embeddings $\instructionTokens_i$ for episode $i$ is randomly classified as correct or wrong. We then predict $J_i$ random token indices, of which each $\hat{\ell}^i_j \in [0, len(\instructionTokens_i)]$, where $J_i$ is the number of errors expected in the dataset (see Tab.~\ref{tab:statistic}).

\noindent\textit{(ii) \CLIPMethod}. For each episode $i$, the set of room and object tokens $\mathcal{K}$ is extracted from instruction $\mathcal{I}$ via an off-the-shelf POS tagger~\cite{nltk}. 
Then, for each observation $O_t$ at time step $t$, the top-$k$ predicted room and object labels are extracted using CLIP~\cite{clip}, forming the set of objects and rooms by observation $\mathcal{S}$. %
The text prompts that are used for finding rooms and objects in the images are composed using the room and object list, as mentioned in Sec.~\ref{sec:task_definition}.
Each CLIP prompt is defined as \textit{``a photo of a: \textless room or object\textgreater"}. 
An instruction is then classified as containing errors if $\mathcal{K} \nsubseteq \mathcal{S}$ (\ie instruction tokens $\mathcal{K}$ are not observed during navigation).
To localize the errors, we retrieve the predicted token indexes as $\{ \hat{l} \,| \, k_{\hat{l}} \in \mathcal{K}: k_{\hat{l}} \notin \mathcal{S} \}$. 

\begin{table*}[ht!]
\caption{Results of our proposed \textit{\oursMethod} method on our proposed benchmark. We show the \texttt{SR} and \texttt{SPL} metrics of the frozen policy, and the drop in \texttt{SR} performance when errors are present ($\Delta$\texttt{SR}$\%$). We then analyze the classification (\texttt{AUC}) and Localization (\texttt{ATD}) performance of different methods. Error types with \commonsense indicate benchmark with common sense. We highlight in gray the main metric, \ie \auc.}
\centering
\resizebox{1\textwidth}{!}{
\begin{tabular}{cc cccc scc scc scc}
\midrule
\multirow{2}{*}{Origin split} & \multirow{2}{*}{Error type} & \multicolumn{2}{m{2.5cm}} {Policy~\cite{an2023bevbert}}&&& \multicolumn{2}{m{2.5cm}}{Random} & & \multicolumn{2}{m{2.5cm}}{\CLIPMethod} && \multicolumn{2}{m{2.5cm}}{\textbf{\oursMethod}} \\ \cmidrule{3-5}\cmidrule{7-8}  \cmidrule{10-11} \cmidrule{13-14}
                              &                             & \scriptsize\textbf{SR}~$\uparrow$ & \scriptsize\textbf{SPL}~$\uparrow$ & \scriptsize\textbf{$\Delta_{SR}(\%)$} && \scriptsize\textbf{AUC}~$\uparrow$ & \scriptsize\textbf{ATD}~$\downarrow$ && \scriptsize\textbf{AUC}~$\uparrow$ & \scriptsize\textbf{ATD}~$\downarrow$ && \scriptsize\textbf{AUC}~$\uparrow$ & \scriptsize\textbf{ATD}~$\downarrow$ &\\ \midrule
\multirow{5}{*}{\makecell{R2R-CE\\Val Unseen}}   
                              & Direction  &0.53 & 0.43 &-18.64&& 0.50&10.54                &&0.50&11.05&      &0.58&8.13 & \\
                              & Room\commonsense      & 0.58& 0.49 &-6.66&& 0.50& 11.03      &&0.57&9.63&   &0.80&7.73& \\
                              & Object\commonsense &0.56 &0.46  &-8.47&&0.51 &10.94          &&0.59&8.76&   &0.74 &9.21 &\\
                              & Room\&Object \commonsense&0.57 & 0.47 &-11.47&& 0.49 &11.56   &&0.64&7.98&  &0.91&7.34& \\
                              & All  \commonsense  &0.52 & 0.43 & -30.64&& 0.51 &12.22         &&0.63&8.68&   &0.94&6.14& \\ \cmidrule{3-14} 
                              & Avg. &0.55&0.46&-15.17    &&0.50&11.26&     &0.59&9.22&    &\textbf{0.79}&\textbf{7.71}\\ \midrule
\end{tabular}}
\label{table:task_1}
\vspace{-0.5cm}
\end{table*}

\textbf{Effect of errors on VLN agent's performance.} This experiment analyzes how each error type affects the performance of a VLN policy in the continuous environment settings in terms of \texttt{SR} and \texttt{SPL}.
Additionally, an analysis of the drop in \texttt{SR} is also computed as follows. 
For each error type, two separate evaluations are performed using only the correct $\mathcal{E_{C}}$ and perturbed $\mathcal{E_{P}}$ episode sets, respectively. Formally, the delta of \texttt{SR} between the two evaluations is defined as $\Delta_{SR}(\%) = SR(\mathcal{E_{C}}) - SR(\mathcal{E_{P}})$. Since $\mathcal{E_{P}}$ is constructed directly from the $\mathcal{E_{C}}$ set, the difference in \texttt{SR} between the two shows the effect of this error type on the policy.
These results are reported in Tab.~\ref{table:task_1} under the \texttt{SR}, \texttt{SPL} and  $\Delta_{SR}(\%)$ columns.
It can be observed that the \textit{Direction} type of error has the largest effect on the navigation policy, with a $-18.64\%$ relative decrease in \texttt{SR}. 
Following that, the \textit{Object} error type with common sense leads to a relative decrease of $-8.47\%$ and the \textit{Room} error with common sense to a decrease of $-6.66\%$. 
Similarly to ~\cite{zhu2022diagnosing}, we find that VLN agents heavily rely on directional and object tokens. Interestingly, we also find that, differently from~\cite{zhu2022diagnosing}, VLN agents in continuous environments are more affected by perturbation of directional tokens than by perturbation of object tokens ($-18.64\%$ \textit{vs} $-8.47\%$). 
We hypothesize that this is due to the ``navigation-graph" of the discrete R2R environment, which may lead to strong implicit assumptions~\cite{krantz_vlnce_2020}. This finding suggests that more focus should be placed on grounding this type of directional information.
Finally, combining the two errors \textit{Room\&Object} with common sense shows a drop in performance of $-11.47\%$, less than the sum of the two separately. 
Lastly, the error type \textit{All} with common sense, which combines all of the above, shows a decrease of $-30.64\%$. 
These results, particularly in the \textit{Direction} case, are a clear sign for the robotic community that efforts need to be made to increase awareness on this weakness in current VLN agents.

\textbf{Does the instruction contain an error?} In this test, the performance of the classification head $f_{d}$ of \textit{\oursMethod} is evaluated against two baselines in terms of \texttt{AUC} scores. The results are reported in Tab.~\ref{table:task_1}.
Here, the \texttt{AUC} scores of a random baseline are presented as a means to identify potential biases and establish a lower bound. 
Given that the ratio between $\mathcal{E_{C}}$ and $\mathcal{E_{P}}$ episodes is $50\%$, \texttt{AUC} for the random baseline is always $\sim0.50$, as expected. 
The \CLIPMethod baseline is presented here as a simple method that reflects how humans would solve the same problem in real life: checking if the content of the instruction is grounded to the observations.
Also, note that \CLIPMethod does not require training.
\CLIPMethod seems to be effective in the presence of \textit{Object} and \textit{Room} types of errors, with a total $0.64$ of \texttt{AUC} score in \textit{Room\&Object} when both errors are present. Interestingly, \CLIPMethod performs on par with the random baseline for the \textit{Direction} error type. 
The intuition is that, when errors are present near the end of the instruction, \CLIPMethod can still ground objects and room tokens to the observations (thus $\mathcal{K} \subseteq \mathcal{S}$), but may subsequently become lost due to the error. 
Finally, in Tab.~\ref{table:task_1}, we show our proposed \textit{\oursMethod} method. We can see that \textit{\oursMethod} achieves the best \texttt{AUC} in all the benchmarks by a large margin. A lower \texttt{AUC} of  $0.58$ for \textit{\oursMethod} also seems to highlight the challenges of the \textit{Direction} error type. Finally, \textit{\oursMethod} has a much higher average \auc score (\textbf{$0.79$}) compared to Random ($0.50$) and \CLIPMethod ($0.59$).

\textbf{Can we localize the error?}
This experiment evaluates the ability of the localization head $f_{l}$ of \textit{\oursMethod} to locate the errors within an instruction. 
The results in terms of \texttt{ATD} for the different baselines are also shown in Tab.~\ref{table:task_1}. 
The random baseline achieves a mean \atd of 11.26. 
The \CLIPMethod baseline performs better for all error types compared to random. 
The policy of considering as token index error $\hat{l}$ the instruction token that was not observed during navigation seems effective, especially in the \textit{Room\&Object} benchmark. 
Finally, the proposed \textit{\oursMethod} achieves the best localization performance across all benchmarks, except for the \textit{Object}, in which \CLIPMethod seems to be particularly effective. 
However, it should be noted that, in the \textit{Object} case, the \auc score for \CLIPMethod is $0.59$ \textit{vs} \textbf{0.74} of \textit{\oursMethod}, which indicates the ability of \CLIPMethod in identifying only a subset of objects that have been swapped.
Finally, it is worth noticing that the mean \atd of \textit{\oursMethod} is $7.71$, which is close to the length of a typical sub-sentence within each instruction, showing that our method can be used to localize errors at the sub-sentence level. 

\textbf{What if we apply \textit{\oursMethod} on the R2R-CE dataset?}
This experiment shows how a pre-trained \textit{\oursMethod} can be used as a semi-automatic tool for potentially identifying error-containing episodes, including those within the original R2R-CE dataset.
Specifically, we apply our \textit{\oursMethod} $f_{d}$ detection head on the original \texttt{Val Unseen} split of R2R-CE.
We then isolate episodes that produce an alignment score $a$ above a set threshold $\tau_a =0.99$.
Surprisingly, \textit{\oursMethod} isolates $25$ episodes out of the total $1839$. Upon manual inspection, we confirmed that $8$ out of the $25$ isolated videos were indeed wrong ground-truth annotations. This experiment suggests the possibility that evaluations in VLN-CE may be skewed due to several wrongly annotated episodes in existing datasets, which should be addressed in future evaluations. Examples can be found in the supplementary materials.

\textbf{Can \textit{\oursMethod} generalize to other datasets and models?} To test this hypothesis, we apply the trained \textit{\oursMethod} on top of another policy, \ie ETPNav~\cite{an2023etpnav}, on the challenging RxR-CE~\cite{rxr}. RxR-CE trajectories and instructions are much longer than R2R-CE (with an average length of 110 tokens \textit{vs} 30 tokens, respectively) and more detailed. Without \textit{any} training on this dataset, \textit{\oursMethod} flagged 118 episodes out of 3669 (we set $\tau_a=0.99$, on the episode with language \texttt{en-IN} and \texttt{en-US}). Surprisingly, after visual inspection 10 of these episodes were found to contain an error, either in a direction, an object, in the goal position, or a mesh issue, and should be removed from the validation split.

\section{Conclusion}

We presented the novel benchmark \datasetName, a new benchmark for VLN-CE, in which different types of errors are injected into textual instructions. Based on our experiments, state-of-the-art VLN-CE methods are affected by instruction perturbations.
On top of \datasetName, we also established the novel task of \textit{\task}. Our proposed method, \textit{\oursMethod}, composed of detection and localization heads, can detect and localize errors within a sub-sentence distance on the original instruction. 
As a further experiment, by running \textit{\oursMethod} on the standard R2R-CE and RxR-CE, we were able to find $8$ and $10$ episodes, respectively, that should be removed
from the validation set. %
Our benchmark and proposed method support the development of more reliable and versatile agents capable of handling real-world ambiguity and uncertainty. 
In future work, we will investigate error-aware policy learning to improve the navigation performance in the VLN-CE task, and online instructions errors detection~\cite{taioli2024i2edlinteractiveinstructionerror}.

\bibliographystyle{IEEEtran}
\bibliography{IEEEabrv,references}

\end{document}